\documentclass[
]{ceurart}
\usepackage[framemethod=TikZ]{mdframed}
\usepackage{arabtex}

\apptocmd{\thebibliography}{\setcounter{enumiv}{6}}{}{}
\usepackage{listings}
\usepackage{url}
\usepackage{graphicx}

\begin{document}

\copyrightyear{2024}
\copyrightclause{Copyright for this paper by its authors.
  Use permitted under Creative Commons License Attribution 4.0
  International (CC BY 4.0).}

\conference{CLEF 2024: Conference and Labs of the Evaluation Forum, September 09–12, 2024, Grenoble, France}

\title{IAI Group at CheckThat! 2024: Transformer Models and Data Augmentation for Checkworthy Claim Detection}

\title[mode=sub]{Notebook for the Checkthat! Lab Task 1 at CLEF 2024}

\author[1]{Peter Røysland Aarnes}[%
orcid=0009-0002-3605-4847,
email=peter.r.aarnes@uis.no
]

\address[1]{University of Stavanger, Kjell Arholms gate 41, 4021 Stavanger, Norway}

\author[1]{Vinay Setty}[%
orcid=0000-0002-9777-6758,
email=vsetty@acm.org,
]

\cormark[1]

\author[1]{Petra Galuščáková}[%
orcid=0000-0001-6328-7131,
email=petra.galuscakova@uis.no,
]

\cortext[1]{Corresponding author.}

\include{text/abstract}


\maketitle

\begin{abstract}
  This paper describes IAI group's participation for automated check-worthiness estimation for claims, within the framework of the 2024 CheckThat! Lab ``Task 1: Check-Worthiness Estimation''. The task involves the automated detection of check-worthy claims in English, Dutch, and Arabic political debates and Twitter data. We utilized various pre-trained generative decoder and encoder transformer models, employing methods such as few-shot chain-of-thought reasoning, fine-tuning, data augmentation, and transfer learning from one language to another. Despite variable success in terms of performance, our models achieved notable placements on the organizer's leaderboard: ninth-best in English, third-best in Dutch, and the top placement in Arabic, utilizing multilingual datasets for enhancing the generalizability of check-worthiness detection. Despite a significant drop in performance on the unlabeled test dataset compared to the development test dataset, our findings contribute to the ongoing efforts in claim detection research, highlighting the challenges and potential of language-specific adaptations in claim verification systems.
\end{abstract}

\section{Introduction}
In an era where information spreads faster than our capacity to verify it, the need for robust mechanisms to assess the veracity of circulating claims has become increasingly critical. In the automated fact-checking research community, a claim is commonly defined as ``an assertion about the world that can be checked'', as formalized by Full Fact \citep{Konstantinovskiy:2021:DT}. However, this definition does not address the worthiness of checking a claim, since not every claim requires scrutiny due to the triviality. To determine whether a claim is checkworthy, several factors could be considered: whether the assertion is of public interest; whether it is factually verifiable, such as statements about the present or the past, or involving correlation and causation; if the claim is a rumor or conspiracy; or if it could potentially cause social harm \cite{Konstantinovskiy:2021:DT, Hassan:2017:SIGKDD, Alam:2021:ACL}. By directing the efforts of fact-checkers and automated systems toward claims with widespread impact, such as those affecting public health or policy decisions, we ensure that critical information remains reliable and verification resources are utilized effectively.

In this paper, we detail our approach to training numerous models for the detection of check-worthy claims, specifically within the framework of the 2024 CheckThat! Lab ``Task 1: Check-Worthiness Estimation'' \citep{Barron:2024:CLEF}. This task seeks to determine whether claims found in tweets or political speech transcriptions merit fact-checking, using a binary classification approach.

We conducted experiments across all three CheckThat! languages chosen by the organizers, English, Dutch, and Arabic. Our submissions ranked best for Arabic, third for Dutch, and ninth for English. We employed various exploratory methods tailored to each language, utilizing various pre-trained autoregressive decoder models and encoder-only transformer models.

For English and Dutch, our primary focus was to fine-tune our chosen models using the training data provided by the organizers for each specific language. However, we also attempted to fine-tune multilingual models using additional data beyond that of the language in which the model would be tested. For Arabic, which would reveal itself being the most challenging dataset, we initially fine-tuned models on Arabic training data. However, the best results were achieved by translating the Arabic test data into English and then using a GPT-3.5 model, fine-tuned in English, to classify the data.

We also took part in Task 2 of the CLEF CheckThat! 2024 challenge, which aimed to determine whether a sentence from a news article expressed the author's subjective viewpoint or presented an objective perspective on the topic. As check-worthy claims are inherently objective statements, we employed the XLM-RoBERTa-Large model, which was trained for claim detection tasks.
Given its multilingual capabilities, we utilized this model for datasets spanning English, German, Italian, Bulgarian, Arabic, and multilingual sources. The XLM-RoBERTa-Large's ability to handle diverse languages made it a suitable choice for this multilingual claim detection task, enabling us to analyze and classify sentences across various linguistic contexts.

\section{Related Work}
As traditional news media experiences a decline in popularity, particularly among younger demographics \citep{Siles:2012:NMS}, platforms like X (formerly known as Twitter) and other types of microblogging services have become primary sources for current events for many individuals. In the influx of X's popularity, the spread of misinformation and fake news has been increasing \citep{Vosoughi:2018:Science}, leading to heightened awareness and concern among researchers, policymakers, and the public. This growing attention has spurred numerous initiatives aimed at combating false narratives, as exemplified by the pervasive misinformation during the 2016 U.S. presidential election \citep{Allcott:2017:JEP, Bovet:2019:Nat} and the COVID-19 \textit{infodemic}, both of which significantly influenced public opinion and health behaviors \citep{Pavlov:2022:MIPRO, Alam:2021:ACL}.

To counteract the spread of misinformation, the research community has intensified efforts to develop datasets and methodologies for automated fact-checking. Claim detection plays a crucial role within these systems, serving as a foundational component for effective automated fact-checking \citep{Guo:2022:TACL}. The most significant progress in this area has been observed in the English language, with the two largest datasets designed for this purpose being ClaimBuster \citep{Arslan:2020:arXiv}, containing approximately 23,500 manually annotated sentences, and CT19-T1 \cite{Atanasova:2019:CLEF}, a dataset being the result of several years' worth of data from the CLEF CheckThat! Lab challenges. 

Additionally, multilingual datasets like those documented by \citet{Gupta:2021:ACL}, primarily used for fact-checking, are also utilized for multilingual claim detection, further enhancing the resources available for this research. Although there exists smaller datasets, typically with fewer than 10,000 annotated sentences, they are predominantly in English \citep{Zeng:2021:arXiv}.

Over the past two years, the CheckThat! Labs have consistently used F1 scores as the official measurement for the check-worthiness estimation subtask. Although the specific task descriptions and the languages tested have varied across different iterations of the CheckThat! Lab, but the overarching goal has remained consistent: to predict the check-worthiness of claims in various languages. This work focuses primarily on text data drawn from sources such as political debates and Twitter \cite{Nakov:2022:CLEF, Alam:2023:CLEF}. For this year's CheckThat! Lab, F1 is again the official measure to assess performance, continuing with a subset of the same languages as previous editions: Arabic, Dutch, and English.

For the 2022 CheckThat! Lab Task 1, focused on check-worthiness estimation, where the NUS-IDS group \citep{du:2022:CEUS} had the winning submission with their CheckthaT5 model, which won in four out of the six language categories that year \citep{Alam:2023:CLEF}. Their model was based on the mT5, a sequence-to-sequence, massively multilingual model \citep{Xue:2021:arXiv}, and was trained jointly on multiple languages to promote language-independent knowledge. Their Arabic submission achieved an F1 score of 0.628, and the Dutch submission had an F1 score of 0.642. The winning English submission, made by the AI Rational group, used a fine-tuned RoBERTa model and achieved an F1 score of 0.698.

For the 2023 CheckThat! Lab Task 1, again focused on check-worthiness estimation (Subtask 1B), the OpenFact group attained the best submission for English. The group fine-tuned GPT-3 which resulted in a F1 of 0.898, and in addition they trained BERT-based model which achieved near identical results \citep{Alam:2023:CLEF, Sawinski:2023:CEUR}. For Arabic, the ES-VRAI group submitted the best results, which were derived from a fine-tuned MARBERT model \citep{Abdul-Mageed:2021:ACL} trained on a downsampled majority class, resulting in a F1 of 0.809 \citep{Sadouk:2023:CLEF}.

\section{Datasets}
\begin{table}
\centering
\caption{Data counts across the training, development, and development test dataset splits.}
\label{tab:data_counts_languages}
\begin{tabular}{lcccccc}
\toprule
Language & Class & Train & Development & Dev-test & Total \\
\midrule
English & No & 17,088 & 794 & 210 & 18,092 \\
 & Yes & 5,413 & 238 & 108 & 5,759 \\
 & Total  & 22,501 & 1,032 & 318 & 23,851 \\[1ex]
\midrule
Dutch & No & 590 & 150 & 350 & 1,090 \\
 & Yes & 405 & 102 & 316 & 823 \\
 & Total & 995 & 252 & 666 & 1,913 \\[1ex]
\midrule
Arabic & No & 5,090 & 682 & 123 & 5,895 \\
 & Yes & 2,243 & 411 & 377 & 3,031 \\
& Total & 7,333 & 1,093 & 500 & 8,926 \\
\bottomrule
\end{tabular}
\end{table}

As shown in Table \ref{tab:data_counts_languages}, there is a significant imbalance in the class label distribution within the training data. If the model is exposed to one class more frequently during training, it may develop a bias towards the majority class, leading to overfitting and poor generalization when encountering the minority class in new data. To address these issues, one can either undersample the majority class or oversample the minority class to create a more balanced training set. Alternatively, other data augmentation techniques such as backtranslations, or synthetic data generation could also be used to balance the class distribution \citep{Henning:2023:EACL}. Additionally, instead of only prioritizing training data class distribution, adjusting the evaluation strategy during training to prioritize maximizing F1 macro-average scores, ensures that predictions for different classes are treated with equal importance.

\section{Methodology}
To conduct our experiments, a series of methods was used in an attempt to optimize the performance of the different fine-tuned models for a given language. These methods include translating data from one language to another to increase the training dataset for the particular given language, text normalization, style transfer, hyperparameter grid searches, and analyzing key performance indicators such as loss and F1 scores during training, which were logged by the Weights \& Biases (W\&B) Python library and online tool \citep{wandb:2020}. In this section, we will explore in greater detail the methods used to fine-tune the different models. The code used to train and test our models is available on our GitHub repository\footnote{IAI group code repository: \url{https://github.com/iai-group/clef2024-checkthat}}.

\subsection{Data pre-processing and augmentation}
Data pre-processing became one of our experimental methods that was used in fine-tuning the different models. We applied the following methods:

\begin{itemize}
    \item{\textbf{Text Normalization: }TweetNormalizer \footnote{TweetNormalizer: \url{https://github.com/VinAIResearch/BERTweet/blob/master/TweetNormalizer.py}} \citep{Nguyen:2020:ACL} script was used post-translation for the Arabic, Dutch and Spanish data. During our preliminary testing, the TweetNormalizer did not yield promising results, leading us to exclude it from further experiments when training our models using hyperparameter grid searches. The reasons behind the poor performance of TweetNormalizer are not entirely clear, although it is plausible that the issue may be related to entity linking. Unlike other approaches, TweetNormalizer does not preserve the specific ``@<username>'' tokens in tweets. Instead, it replaces any distinct username with a generic ``@USER'' token, effectively removing unique identifiers associated with different classes. This removal of specific usernames could potentially disrupt contextual relevance, which might otherwise contribute positively when fine-tuning the models.} 
    \item{\textbf{Machine Translation: }
    Due to the large amounts of data to translate, we opted to use a free of charge translation systems available in the deep translator library. According to the recent WMT report \citep{Kocmi:2023:WMT}, the quality of such freely available commercial systems depends on the particular language pair, but is relatively high for all the studied systems and language pairs. We thus used Google Translate implementation from deep-translator\footnote{Google Translate deep translator: \url{https://deep-translator.readthedocs.io/en/latest/usage.html\#google-translate}} due to its support of all studied languages, usual performance quality and no required subscription or API key. Google Translate was used to translate datasets from any provided source language (English, Dutch, Arabic, and Spanish) to any target language (English, Dutch, Arabic).}
    \item{\textbf{Style Transfer: } As the style of the English collection (political debates) substantially differs from the style of Dutch and Arabic collections (Twitter data), we also experimented with machine translation with style transfer to prepare in-style training data. Specifically, we style transferred the translated English training data to resemble a closer match to the Arabic data. To do this style transfer, we employed gpt-3.5-turbo-0125 model via ChatGPT API. We use a single prompt for each sentence in which we ask the system to translate the sentence and also to transfer the style of debate into a Tweet.  
    We used a few-shot approach with three example Tweets selected from the Arabic training collection and the following prompt: \textit{Rephrase the following statement as if somebody was Tweeting about it in Arabic. Output might use hashtags, emoticons, images and links. Statement:  (\{text to translate\}) + Here are a few examples: (\{arabic examples\})}.
    Though the quality of the translated sentences looked reasonable, using these data did not lead to any improvement, suggesting that the domain mismatch between the collections is too large to be crossed just by a style change. Style transfer might even affect the check-worthiness of the claim. GPT model used for style transfer was paid and also relatively slow, what did not allow more extensive experimentation.}   
\end{itemize}

\textbf{Few-shot chain-of-thought reasoning instruction prompt}
\begin{mdframed}[backgroundcolor=gray!20, linecolor=black, linewidth=1pt, innerleftmargin=10pt, innerrightmargin=10pt, innertopmargin=10pt, innerbottommargin=10pt]
\texttt{Your task is to identify whether a given tweet text in the \{lang\} language is verifiable using a search engine in the context of fact-checking.\\
Let's define a function named checkworthy(input: str).\\
he return value should be a strings, where each string selects from "Yes", "No".\\ \\
"Yes" means the text is a factual checkworthy statement.\\
"No" means that the text is not checkworthy, it might be an opinion, a question, or others.\\
For example, if a user call checkworthy("I think Apple is a good company.")\\
You should return a string "No" without any other words, \\
checkworthy("Apple's CEO is Tim Cook.") should return "Yes" since it is verifiable.\\ \\
Note that your response will be passed to the python interpreter, SO NO OTHER WORDS!\\
Always return "Yes" or "No" without any other words.\\
\
checkworthy(\{text\})}
\label{prompt}
\small
\end{mdframed}

\subsection{Model Types and Fine-tuning}
\label{subsec:model_types}
For our experiments, we utilized both pre-trained generative autoregressive decoder transformer models and pre-trained encoder-only transformer models to assess their effectiveness in predicting text in English, Dutch, and Arabic. Our selection of generative models was based on their popularity and availability, which includes GPT-4 \citep{OpenAI:2024:arXiv}, Mistral-7b \citep{Jiang:2023:arXiv}, GPT-3.5 with few-shot chain-of-thought (CoT) reasoning, and a fine-tuned GPT-3.5 \citep{Ouyang:2022:arXiv}. For the encoder models, we chose XLM-RoBERTa-Large \citep{Conneau:2020:arXiv} and RoBERTa-Large \citep{Liu:2019:arXiv}, which are prominent in multilingual training classification and English classification tasks, respectively.

For fine-tuning the encoder-only models, we utilized the Hugging Face Trainer\footnote{\url{https://huggingface.co/docs/transformers/main_classes/trainer}} class. Although most hyperparameters were kept in their default settings, the number of epochs was set to a static 50. The development dataset was evaluated after each epoch, optimizing for Macro F1 score to monitor performance. We also employed the hyperparameter grid search using Weights \& Biases \citep{wandb:2020} sweep functionality to conduct multiple training runs, testing the most critical hyperparameter combinations. In an attempt to save time during training, training would terminate early if the F1 score for the development dataset did not improve after 3 consecutive epochs. 

The following list contains an overview of the different models in our experiments used, including what data was used for fine-tuning, and specifies if a particular model was only used for one particular language.

\begin{itemize}
    \item \textbf{GPT-4} \citep{OpenAI:2024:arXiv}: Few-shot CoT reasoning. Tested on all three languages.
    \item \textbf{Mistral 7b }\citep{Jiang:2023:arXiv}: Few-shot CoT reasoning. Tested on all three languages.
    \item \textbf{GPT-3.5} \citep{Ouyang:2022:arXiv}: Few-shot chain-of-thought reasoning approach. Tested on all three languages.
    \item \textbf{GPT-3.5} \citep{Ouyang:2022:arXiv} (fine-tuned): Fine-tuned on English training data for the English tests and Arabic to English translations, and another model was fine-tuned on Spanish, Arabic and Dutch for the Dutch test.
    \item \textbf{XLM-RoBERTa-Large} \citep{Conneau:2020:arXiv} (XLMR): Fine-tuned on English ClaimBuster \cite{Arslan:2020:arXiv}, Norwegian and German podcasts data for claim detection \cite{becker:2023:uis}. The model was tested on all three languages.
     \item \textbf{XLM-RoBERTa-Large (fine-tuned)} \citep{Conneau:2020:arXiv}: This version of XLM-RoBERTa (which we will refer to as ``XLMR fine-tuned'') builds upon the initial fine-tuning of the aforementioned XLMR model. It underwent additional fine-tuning with the organizer's training data, specifically tailored for a particular language. It was evaluated across all three languages.
    \item \textbf{RoBERTa-Large} \cite{Liu:2019:arXiv}: Fine-tuned on unaltered English organizer's training data, and was tested only on English data.
\end{itemize}

\subsubsection{English Model fine-tuning and hyperparameter tuning}

\begin{figure}[htbp]
  \centering
    \includegraphics[scale=0.185]{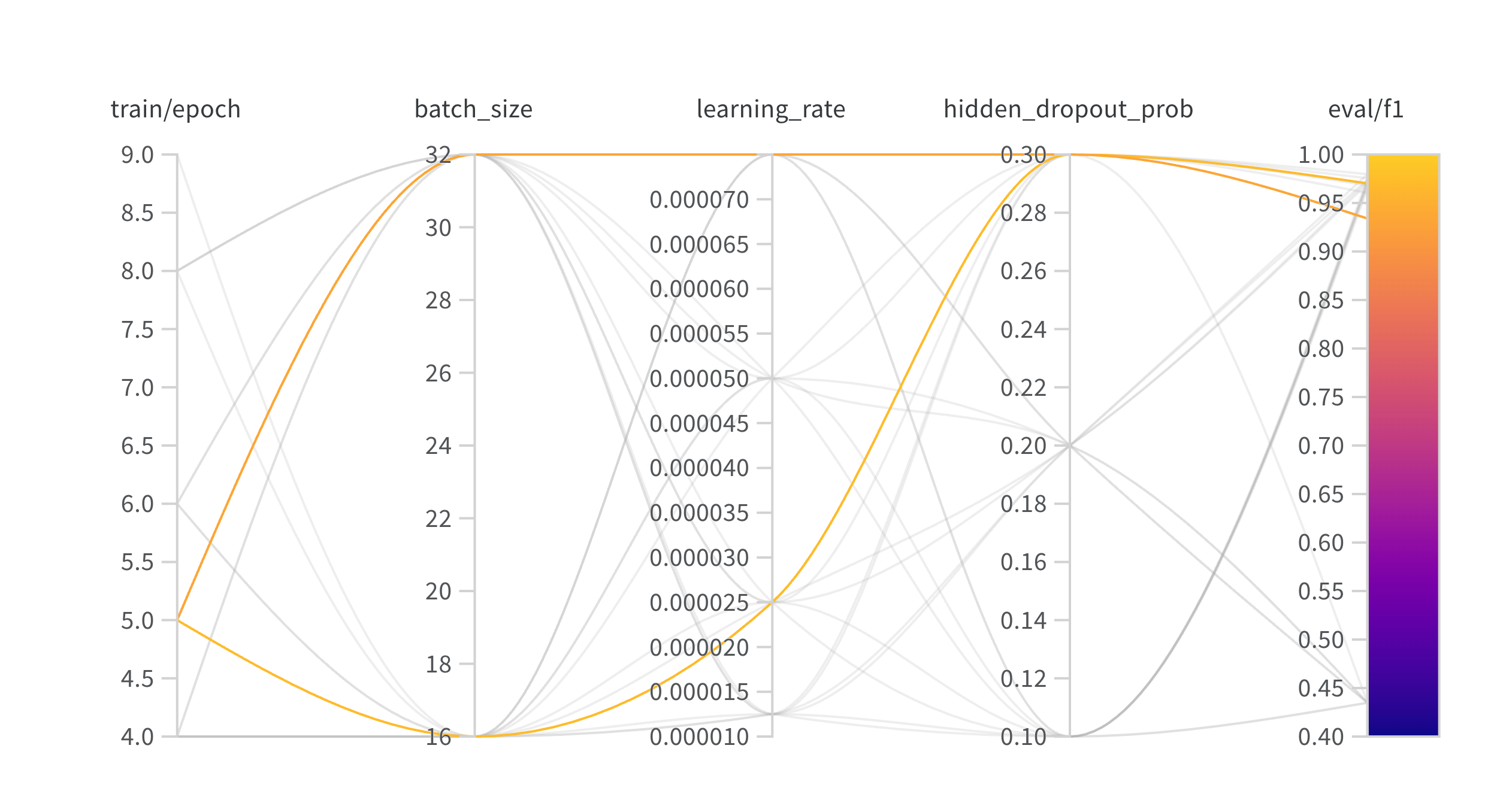}
  \caption{Hyperparameter W\&B parallel coordinates plot for RoBERTa hyperparameter grid search. Eval/f1 relates to the best dev-test F1 in a given run.}
  \label{wb_roberta}
\end{figure}

For the organizer English unlabeled test data submission, we used a fine-tuned RoBERTa-Large model, since it outperformed the other models on the development test (dev-test) datasets. The list of hyperparameters employed for the grid search is provided in Table \ref{tab:hyperparameters}. Figure \ref{wb_roberta} illustrates the outcome of the 24 distinct training runs and their corresponding performance on the development dataset.

\begin{table}[h!]
\centering
\caption{Hyperparameters grid search values for RoBERTa-Large English fine-tuning. *Early termination if F1 score did not improve after 3 consecutive epochs.}
\begin{tabular}{ll}
\textbf{Parameters}       & \textbf{Values}                 \\ \hline
Batch Size                & 16, 32                          \\
Epochs                    & 50*                             \\
Hidden Dropout Probability & 0.1, 0.2, 0.3                  \\
Learning Rate             & 1.25e-05, 2.5e-05, 5e-05, 7.5e-05 \\ \hline
\end{tabular}
\label{tab:hyperparameters}
\end{table}

Given the consistently high F1 scores across multiple RoBERTa training runs on the development and dev-test datasets, more in depth analysis was conducted. During the RoBERTa grid search, two specific model runs, which we will refer to as model $A$ and model $B$, performed exceptionally well on the development and dev-test datasets. Since only one model prediction results could be submitted for the final evaluation, our objective was to make an informed decision on which model would likely perform the best. For example, model $A$ demonstrated slightly better dev-test F1 scores compared to model $B$, although model $B$ performed better than model $A$ on the development during $B$'s training.

As a final sanity check, we compared the prediction overlap between models $A$ and $B$ and a fine-tuned GPT-3.5 model to determine which RoBERTa model deviated most from the GPT-3.5's predictions. A significant difference in overlap with the GPT model suggests that one of the RoBERTa models might have developed a unique pattern of predictions, which in turn could have a significant impact on its performance on real-world data. We hypothesized that a higher percentage of overlap in predictions with the GPT model would be advantageous.

Based on comparisons and analysis of key performance indicators, such as the development dataset F1, dev-test F1, and the loss rate, we systematically gathered and analyzed training and testing data using W\&B, including test data prediction overlaps. As a result, we decided to go with the RoBERTa model $A$.

\subsubsection{Dutch Model fine-tuning}
For Dutch, we utilized XLMR, which was fine-tuned on ClaimBuster and podcast data (as detailed in Section \ref{subsec:model_types}), as well as XLMR fine-tuned with datasets in the four languages provided by the organizers. Additionally, we used GPT-3.5 fine-tuned on Dutch, Arabic, and Spanish data, and finally, leveraged LLMs GPT-4 and Mistral-7b with few-shot CoT reasoning prompts. After extensive analysis, for Dutch, GPT-4 was the best performing model on the dev-test dataset.

\subsubsection{Arabic Model fine-tuning}
For Arabic, none of the XLMR models or LLMs with CoT prompt performed well. Since we suspected that the distribution of dev-test and organizer test are different, we randomly sampled 10\% of the test dataset and manually annotated it with labels, thereafter tested each model on that annotated sample. Since we are three contributors to these experiments, each person annotated labeled 10\% of the sample separately. We calculated Cohen’s kappa to assess inter-annotation agreement ($k=0.424$). In cases of inter-annotation disagreement, the sentence in question would be annotated according to the majority rule.

\section{Results and Discussion}
In this section, we present the results that the different models produced for the dev-test and the submission test dataset after the gold standard got published. For the dev-test set we access the performance using metrics that include accuracy, precision, recall, and F1 scores for the positive class (check-worthy claims), moreover, for the test dataset, only the F1 was measured.

\subsection{English}
\begin{table}
\centering
\caption{English performance binary averages metrics for the ``check-worthy''-class. The columns for accuracy, precision, and recall are measured from the dev-test dataset.}
\label{tab:EN_results}
\begin{tabular}{lrrrr|r}
\toprule
Model &  Accuracy &  Precision &  Recall &  F1 (dev-test) & F1 Test  \\
\midrule
GPT-4 &     0.808 &     0.813 &  0.565 &               0.667 &               0.658 \\
Mistral-7b &     0.726 &     0.667 &  0.389 &               0.491 &               0.503 \\
GPT-3.5 &     0.745 &     0.865 &  0.296 &               0.441 &               0.397 \\
GPT-3.5 (fine-tuned) & 0.915  & \textbf{0.966}  &  0.778  & 0.862  & 0.705 \\
XLMR &     0.830 &     0.829 &  0.630 &               0.716 &               0.717 \\
XLMR (fine-tuned) &     0.767 &     1.000 &  0.315 &               0.479 &           0.662 \\
RoBERTa &     \textbf{0.937} &     0.958 &  \textbf{0.852} &               \textbf{0.902} &              \textbf{0.753} \\
\bottomrule
\end{tabular}
\end{table}

The models evaluated in English includes GPT-4, Mistral-7b, GPT-3.5, XLMR, XLMR (fine-tuned), and RoBERTa. Table \ref{tab:EN_results} provides a detailed overview of the metrics for the positive class.

\begin{itemize}
\item RoBERTa emerged as the best performing model for accuracy (0.937), precision (0.958), and recall (0.852) for the dev-test data, reflecting a strong ability to correctly identify relevant instances without a high rate of false positives. This resulted in an impressive F1 score of 0.902 for the dev-test, however, the F1 decreased by 0.099 for the test data (F1 0.753). Conversely, Mistral-7b and GPT-3.5 showed lower performance across most metrics, with Mistral-7b demonstrating a particular weakness in precision (0.667), and GPT-3.5 with even worse recall (0.296).

\item GPT-4 and XLMR displayed moderate performance, with XLMR having a slight edge over GPT-4 in accuracy and F1 scores. Interestingly, the fine-tuned XLMR (fine-tuned) achieved a perfect precision score of 1.000 but at the cost of lower recall (0.315), suggesting a conservative prediction behavior that limited its false positives, but missed several relevant predictions.

\item The variation in performance across the test and dev-test dataset for our best performing model, RoBERTa, suggests potential overfitting or dataset-specific biases which makes it poor at generalizing across different data. Efforts to that could be beneficial for future experiments would be fine-tuning with an expanded parameter grid search, data augmentation such as testing oversampling or undersampling techniques, or using additional translated English data to make the training data more diverse which could potentially help the model's ability to generalize.

\end{itemize}

\subsection{Dutch}

The models evaluated for the Dutch language include GPT-4, Mistral 7b, GPT-3.5, XLMR, and XLMR (fine-tuned). Table \ref{tab:dutch_results} provides a detailed overview of the performance metrics for the different models. This condensed analysis aims to highlight which models perform best in handling Dutch language data, emphasizing their strengths and potential areas for improvement. For the final submission, GPT-4 was the model used.

\begin{itemize}
\item Overall performance for Dutch, XLMR (fine-tuned) demonstrated the best F1 scores, for the dev-test data (0.653), with a slight performance decrease for test data (0.611). This model excelled particularly in recall (0.722) compared to the other models. The high recall coupled with reasonable precision (0.597) suggests a balanced approach to maximizing both positive identifications and accuracy of predictions. 

\item GPT-4, Mistral 7b, and GPT-3.5 (all three using CoT reasoning) showed weaker performance metrics overall compared to XLMR (fine-tuned). GPT-4, despite GPT-4's lower accuracy and precision in the dev-test (0.577 and 0.580, respectively), showed a significant increase in F1 score on the test data (0.718), which may indicate better generalization under specific conditions. Mistral 7b, on the other hand, displayed lower metrics across the board with particularly low recall (0.310).

\item The XLMR model, while not reaching the heights of its fine-tuned counterpart on the dev-test data, still outperformed the GPT models in most metrics on the dev-test dataset, showing particular strength in recall (0.611) that closely matches its precision (0.603). This balance resulted in robust F1 scores in both the dev-test (0.607) and test (0.694) scenarios, underlining its utility as a reliable model for this task.

\item All the models showed significant performance variances across the datasets. Interestingly, only the XLMR (fine-tuned) exhibited a performance decline from the dev-test to the test dataset, while all other models performed significantly better on the test dataset. Notably, the GPT-3.5 model, fine-tuned on Spanish, Arabic, and Dutch, achieved an F1 score of 0.781 on the test dataset. This score would have placed it at the top of the CheckThat! leaderboard for Dutch, had we submitted these results instead of those from GPT-4.

\end{itemize}

\begin{table}
\centering
\caption{Dutch performance binary averages metrics for the ``check-worthy''-class. The columns for accuracy, precision, and recall are measured from the dev-test dataset.}
\label{tab:dutch_results}
\begin{tabular}{lrrrr|r}
\toprule
Model &  Accuracy &  Precision &  Recall &  F1 dev-test & F1 Test \\
\midrule
GPT-4 &     0.577 &     0.580 &  0.389 &               0.466 & 0.718 \\
Mistral 7b &     0.547 &     0.539 &  0.310 &               0.394 & 0.601 \\
GPT-3.5 &     0.538 &     0.544 &  0.155 &          0.241 & 0.647 \\
GPT-3.5 (fine-tuned) &   0.677  &  0.706  &  0.547   &  0.617  & \textbf{0.781}  \\
XLMR &     0.625 &     \textbf{0.603} &  0.611 &         0.607 & 0.694 \\
XLMR (fine-tuned) &     \textbf{0.637} &     0.597 & \textbf{ 0.722} &               \textbf{0.653} & 0.611 \\
\bottomrule
\end{tabular}
\end{table}

\subsection{Arabic}
\begin{table}
\centering
\caption{Arabic performance binary averages metrics for the ``check-worthy''-class. The columns for accuracy, precision, and recall are measured from the dev-test dataset.}
\label{tab:AR_results}
\begin{tabular}{lrrrr|r}
\toprule
Model &  Accuracy &  Precision &  Recall &  F1 dev-test & F1 Test  \\
\midrule
GPT-4 &     0.810 &     \textbf{0.890} &  0.854 &      0.871 &     0.526 \\
Mistral 7b &     0.700 &     0.865 &  0.714 &   0.782 &   0.493 \\
GPT-3.5 &     0.664 &     0.862 &  0.661 &   0.799 &  0.397 \\
GPT-3.5 (fine-tuned) &     \textbf{0.824} &     0.885 &  \textbf{0.881} &  \textbf{0.883} &   \textbf{0.569} \\
XLMR &     0.784 &     0.848 &  0.870 &  0.859 &  0.549 \\
XLMR (fine-tuned) &     0.740 &     0.919 &  0.719 &  0.807 &  0.519 \\
\bottomrule
\end{tabular}
\end{table}

The models evaluated for the Arabic language include GPT-4, Mistral 7b, GPT-3.5, XLMR, and XLMR (fine-tuned). In addition, a fine-tuned English GPT-3.5 model was evaluated, which classifies Arabic-to-English translated data. Table \ref{tab:AR_results} provides a detailed overview of the performance metrics for the ``check-worthy'' class, encompassing accuracy, precision, recall, and F1 for the dev-test dataset, as well as the F1 score for the submission test dataset. In addition, we annotated 10\% of the test data prior to getting the gold standard, attempting to gain a greater understanding of how the different models might behave for the test data. The most promising model tested on the 10\% sample data was the fine-tuned GPT-3.5, which attained an F1 score of 0.848. Consequently, we chose this model for our final submission.

\begin{itemize}

\item GPT-3.5, fine-tuned on English, outperformed the GPT-4 CoT (except for precision), it also outperformed GPT-4 on the 10\% annotated sample data. As a result of this, we chose to submit test results from the GPT-3.5 model. However, there was a significant drop in performance on the test data, where the F1 score decreased to 0.569. This indicates a potential issue with the model's ability to generalize from the development environment to more diverse or challenging test scenarios. 

\item GPT-4 with CoT training, demonstrated robust performance across most metrics, achieving the highest precision in the dev-test set (0.890) and showcasing strong F1 score (0.871) and recall (0.854). However, it underperformed compared to GPT-3.5 fine-tuned. We see a similar drop in performance on test set, indicating that it is significantly different from dev-test.

\item XLMR showed consistent performance, with particularly great recall (0.870) on the dev-test, translating into an F1 score of 0.859. It attained the second highest F1 score on the standard test dataset (0.549)

\item The XLMR (fine-tuned) also performed well, improving on precision (0.919) significantly compared to XLMR counterpart on the dev-test data, which resulted in an F1 score of 0.807. However, like GPT-4, as for all other models as well, XLMR (fined-tuned) saw a decrease in performance on the test dataset (F1 0.519), which could suggest an overfitting to the dev-test environment or a need for further tuning to enhance its ability to generalize across different data.

\end{itemize}

\section{Conclusion and Future Work}
This study offers a detailed examination of the 2024 CheckThat! Lab competition, Task 1, focusing on check-worthiness estimation for claims in political debates and Twitter data in English, Dutch, and Arabic. We employ a strategic combination of few-shot chain-of-thought reasoning and language-specific fine-tuning methods.

Our submissions attained the first place for Arabic with an F1 of 0.569, where we translated the Arabic test data to English, thereafter used a fine-tuned GPT-3.5 for English to classify the translated data. For Dutch, we secured the third-best submission with a F1 of 0.718, using GPT-4 with few-shot chain-of-thought reasoning. Lastly, for the English submission earned us the ninth-best submission with the F1 score of 0.753 using a RoBERTa-Large model, trained on unaltered English training data provided by the competition organizers.

Despite having the best submission for Arabic, we observed a significant drop in performance when comparing the results from the dev-test and the actual submission test dataset. This signals possible challenges such as overfitting and poor generalization across unseen data. These issues would be an important area for future investigations, possibly through more robust model training techniques and exploring additional data augmentation strategies.

\section{Acknowledgments}
This research is funded by SFI MediaFutures partners and the Research Council of Norway (grant number 309339).

\bibliography{CLEF2024}

\end{document}